\documentclass[hidelinks]{article}

\usepackage[final, nonatbib]{neurips_2023}

\usepackage[utf8]{inputenc} 
\usepackage[T1]{fontenc}    
\usepackage{hyperref}       
\usepackage{url}            
\usepackage{booktabs}       
\usepackage{amsfonts}       
\usepackage{nicefrac}       
\usepackage{microtype}      
\usepackage{xcolor}         

\usepackage[style=numeric]{biblatex}
\bibliography{bibliography}
\usepackage{graphicx}

\title{RainAI - Precipitation Nowcasting from Satellite Data}

\author{%
  Rafael Pablos Sarabia \\
  Aarhus University \\
  Department of Computer Science \\
  Cordulus \\
  \texttt{rpablos@cs.au.dk} \\
  \And
  Joachim Nyborg \\
  Cordulus \\
  \texttt{jn@cordulus.com} \\
  \AND
  Morten Birk \\
  Cordulus \\
  \texttt{mb@cordulus.com} \\
  \And
  Ira Assent \\
  Aarhus University \\
  Department of Computer Science \\
  DIGIT Aarhus University Centre for \\
  Digitalisation, Big Data and Data Analytics \\
  \texttt{ira@cs.au.dk} \\
}

\begin{document}

\maketitle

\begin{abstract}
This paper presents a solution to the Weather4Cast 2023 competition, where the goal is to forecast high-resolution precipitation with an 8-hour lead time using lower-resolution satellite radiance images. We propose a simple, yet effective method for spatiotemporal feature learning using a 2D U-Net model, that outperforms the official 3D U-Net baseline in both performance and efficiency. We place emphasis on refining the dataset, through importance sampling and dataset preparation, and show that such techniques have a significant impact on performance. We further study an alternative cross-entropy loss function that improves performance over the standard mean squared error loss, while also enabling models to produce probabilistic outputs. Additional techniques are explored regarding the generation of predictions at different lead times, specifically through Conditioning Lead Time. Lastly, to generate high-resolution forecasts, we evaluate standard and learned upsampling methods. The code and trained parameters are available at \url{https://github.com/rafapablos/w4c23-rainai}.
\end{abstract}

\section{Introduction}
Short-term rainfall prediction, also known as precipitation nowcasting, plays a crucial role in various industries, such as agriculture, energy management, emergency services, and transportation. The objective of precipitation nowcasting is to generate accurate and high-resolution forecasts of rainfall intensity within a specific geographic area over a short time span ranging from 1 to 24 hours.

Traditional methods in this field rely on Numerical Weather Prediction (NWP) and optical flow techniques, which are based on the physics of the atmosphere and mathematical models \cite{nwp_guidelines, rainymotion, pysteps}. However, the slow runtime of NWP models make them unsuitable for short-term nowcasting, and optical flow methods are constrained by their assumption of constant motion for precipitation features over time \cite{agrawal_machine_2019, ravuri_skilful_2021}. To address these limitations, there has been a shift towards employing data-driven methods based on deep learning. These models learn from historical observations and exploit the processing speed of modern Graphics Processing Units (GPUs) to generate forecasts faster than NWPs and capture more intricate patterns than optical-flow methods due to their non-linear nature. However, the large spatial and temporal context required for nowcasting presents a significant challenge for deep learning methods, which must find a balance between efficient processing and capturing adequate context across high spatial and temporal dimensions.

The Weather4Cast 2023 Competition on Super-Resolution Rain Movie Prediction under Spatio-temporal Shifts at the 37th Conference on Neural Information Processing Systems (NeurIPS 2023) provides an opportunity to work on the intricate task of precipitation nowcasting. Similar to the Weather4Cast 2022 competition \cite{gruca_weather4cast_2022}, this competition focuses on forecasting future high-resolution rainfall events using lower resolution satellite radiances. Successful rainfall prediction in this context requires data fusion, multi-channel video frame prediction, and super-resolution techniques. Notably, a new challenge in this year's competition involves the quantitative prediction of rainfall, contrasting with the previous year's focus on determining whether rain would surpass a predefined threshold.

Our work explores a simple architecture based on the 2D U-Net \cite{ronneberger_u-net_2015} as the foundational structure and employs different encoders — specifically ResNet-18 \cite{resnet} blocks or Swin Transformer \cite{liu_swin_2021}, detailed in Sections \ref{s:unet} and \ref{s:swin} respectively. This is due to the ability of the 2D U-Net to effectively process spatial dimensions in the input data. The alternative encoders are meant to enhance the original 2D U-Net to capture more complex and long-term dependencies. Emphasis is placed on optimizing the dataset by applying sampling and problem-specific modifications as outlined in Section \ref{s:data}. To enhance training, we adopt a classification approach with the cross-entropy loss function, elaborated on in Section \ref{s:loss}. Another proposal consists of extending the model to condition the output to the required lead time instead of generating all timesteps simultaneously (Section \ref{s:clt}). Additionally, we incorporate super-resolution techniques to derive high-resolution predictions from the central region of the model's output, explained in Section \ref{s:upsample}.

The models we propose surpass the official baseline across the core, nowcasting, and transfer-learning leaderboards, as measured by the average Critical Success Index (CSI) at various intensity thresholds. In comparison to other submissions, our best models rank within the top 5 teams on each leaderboard.

\section{Related Work}
We classify existing efforts in nowcasting into two paradigms: traditional and deep learning-based. The fundamental distinction between traditional and deep learning methods is that the former rely on physical models while the latter models learn from historical weather data including radar maps, satellite images, and recorded meteorological variables such as previous precipitation or wind.

The traditional approaches include Numerical Weather Prediction (NWP) models and optical flow models such as RainyMotion \cite{rainymotion} and PySteps \cite{pysteps}. NWP models have been established to predict the weather based on mathematical models \cite{nwp_guidelines}, but they are dependent on initial and boundary conditions at the top, lateral, and lower boundaries of the model domain. In order to obtain probabilistic and better forecasts, ensemble NWP systems are used to simulate coupled physical equations of the atmosphere with different perturbations and initial conditions \cite{ensemble_nwp}. With respect to optical-flow methods, they assume that the movement of radar echoes along a fixed motion field can capture the evolution of precipitation for a few hours without any changes in intensity. Therefore, they estimate the motion field and extrapolate radar echoes to create a forecast. These forecasts can also be combined with ensembles to generate more accurate probabilistic forecasts. These traditional approaches for precipitation nowcasting based on physical models require high computing power, especially for ensemble models, so the spatial and temporal resolution tends to be limited. Additionally, due to their convergence time, they are not the most adequate for precipitation nowcasting, where the primary interest is in the first few hours of the forecast. With respect to extrapolation-based approaches, they tend to overestimate precipitation and not cover the region with rain properly \cite{ravuri_skilful_2021}.

Due to the effectiveness of deep learning in a wide range of tasks, particularly in computer vision and natural language processing, there has been a growing interest in applying deep learning to precipitation nowcasting. Deep learning models have demonstrated improved forecast quality, as measured by per-grid-cell metrics, due to their direct optimization and fewer inductive biases. Deep learning methods leverage modern Graphical Processing Units (GPU) to generate forecasts in seconds \cite{metnet2} and can capture complex non-linear precipitation phenomena due to their ability to identify patterns in high-dimensional data \cite{goodfellow_deep_2016}.

A number of deep learning approaches for precipitation nowcasting are based on the U-Net architecture \cite{ronneberger_u-net_2015} consisting of an encoder-decoder architecture with Convolutional Neural Network (CNN) layers. While this architecture has no explicit modeling of memory and simply manages the temporal dimension as an additional channel, multiple precipitation nowcasting models based on the original U-Net architecture have been successfully employed \cite{agrawal_machine_2019, trebing_smaat-unet_2021, ayzel_rainnet_2020, fernandez_broad-unet_2021}. Additional architectures, including a variety of recurrent neural networks (RNNs), have been proposed with a dedicated mechanism to handle temporal information and fully capture temporal dynamics and dependencies \cite{shi_convolutional_2015, shi_deep_2017, luo_pfst-lstm_2021}. The main challenge with RNNs is training time due to its recurrent connections and unrolling of the time sequences.

Additional advanced deep learning techniques used in weather forecasting include MetNet \cite{metnet}, MetNet2 \cite{metnet2} and MetNet3 \cite{metnet3} to forecast precipitation in the United States for up to 8, 12, and 24 hours respectively. While the resources required for these models are extremely large, many of the insights and decisions can be transferred to smaller models with less parameters and inputs with lower dimensionality. Some of these insights cover data preprocessing and transformations, problem formulation, architecture choices, training optimization, and evaluation metrics. \cite{bi_pangu-weather_2022, ravuri_skilful_2021, bi_accurate_2023, zhang_skilful_2023, pihrt_weatherfusionnet_2022} also make use of large models with complex architectures and training including Vision Transformers \cite{dosovitskiy_image_2021}, Swin Transformers \cite{liu_swin_2021}, MaxVit \cite{tu_maxvit_2022}, PhyDNet \cite{guen_disentangling_2020}, Earthformer \cite{gao_earthformer_2022}, and Generative Adversarial Training \cite{goodfellow_generative_2014}. A main weakness of most of these models is their complexity and required computing power when compared to more simple architectures such as the 2D U-Net.

\section{Weather4Cast Challenge \& Dataset}
\label{competition}
The task for the competition involves predicting 32 ground-radar images that quantitatively represent the amount of rainfall for the next 8 hours, given an input sequence of satellite observations for the preceding hour (4 time slots). The organizers provide satellite and radar images for 7 regions in Europe with different precipitation characteristics, spanning from February 2019 to December 2020. Additionally, for evaluating spatial and temporal transfer learning, 3 more regions and data for 2021 are provided. For precipitation nowcasting, radar data is more useful and has higher accuracy and resolution than satellite data. However, satellite data is used as input to the models in this competition because radar observations are expensive to obtain and not available in many parts of the world.

The input satellite data is obtained by geostationary satellites operated by the European Organisation for the Exploitation of Meteorological Satellites (EUMETSAT) with a spatial resolution of about 12 x 12 km$^2$ and a temporal resolution of 15 minutes. This data contains 11 channels, including visible and near-infrared (VIS), thermal infrared (IR), and water vapor absorption (WV) bands for the selected regions. Given the input patch has size 252 x 252 pixels, the input data covers an area of about 3,024 x 3,024 km$^2$ for each region.

For each input region of about 3,024 x 3,024 km$^2$, the objective is to forecast precipitation in the central area of the region. In order to evaluate rainfall predictions in each region, rainfall rates, derived from ground-radar reflectivity, are provided by the Operational Program for Exchange of Weather Radar Information (OPERA) radar network. This radar data is provided in 15-minute intervals in patches of size 252 x 252 pixels with a spatial resolution of about 2 x 2 km$^2$, covering the central area of about 504x504 km$^2$ for each region.

The challenge involves predicting precipitation rates in a region of size 504 x 504 km$^2$, corresponding to the central region of a much larger area covering about 3,024 x 3,024 km$^2$ for which satellite measurements are available. This implies about 1,260 km of spatial context in each direction. Considering the required 64-85 km of context per hour of lead time in each spatial dimension \cite{metnet2}, enough spatial context is provided to generate predictions for up to 15-19 hours.

Regarding the available data for the 7 regions and 2 years used for training, there are 228,928 training sequences generated by the provided code with a sliding window, and 840 predefined sequences for validation. While this is a large amount of training data, the data is highly unbalanced towards no rain, which makes it difficult to train an unbiased model. Table~\ref{val-data-distribution} shows the data distribution of the validation data for different rainfall intensities using one of the proposed classification scales from PySTEPS \cite{pysteps}. Note that not only do lower rainfall intensities have many more events, but they also cover smaller ranges.

\begin{table}[h]
  \caption{Rainfall distribution over the validation dataset (7 regions, 2019-2020)}
  \label{val-data-distribution}
  \centering
  \begin{tabular}{ll}
    \toprule
    Intensity (mm/h) & Observations (\%) \\
    \midrule
    < 0.08 & 88.05\\
    (0.08, 0.16] & 3.57\\
    (0.16, 0.25] & 1.71\\
    (0.25, 0.40] & 1.60\\
    (0.40, 0.63] & 1.53\\
    (0.63, 1.0] & 1.34\\
    (1.0, 1.6] & 1.00\\
    (1.6, 2.5] & 0.61\\
    \bottomrule
  \end{tabular}
  \quad
  \begin{tabular}{ll}
    \toprule
    Intensity (mm/h) & Observations (\%) \\
    \midrule
    (2.5, 4.0] & 0.35\\
    (4.0, 6.3] & 0.15\\
    (6.3, 10.0] & 0.06\\
    (10.0, 16.0] & 0.02\\
    (16.0, 25.0] & 0.01\\
    (25.0, 40.0] & 0.00\\
    > 40.0 & 0.00\\
    \\
    \bottomrule
  \end{tabular}
\end{table}

In addition, for each of the regions, static data with latitude, longitude, and topological height is provided.

\section{Model Architecture \& Optimizations}
The contributions for the competition, detailed in this section, focus on different architectures, optimization of the dataset, and changes to the problem formulation. These changes include approaching the problem as classification with cross-entropy loss, integration of lead time conditioning, and the incorporation of super-resolution techniques to obtain the final high-resolution precipitation forecasts.

\subsection{2D U-Net}
Given its success in various tasks, including precipitation nowcasting \cite{agrawal_machine_2019, ayzel_rainnet_2020, fernandez_broad-unet_2021, trebing_smaat-unet_2021}, the original 2D U-Net architecture \cite{ronneberger_u-net_2015} is studied. For this challenge, modifications are made to the original U-Net by incorporating Residual Networks (ResNet) \cite{resnet} blocks or Swin Transformer \cite{liu_swin_2021} as the encoder. 

The U-Net is a fully-convolutional architecture consisting of an encoder to generate feature maps containing high-level information and a decoder to obtain full-resolution outputs. The use of skip connections is essential to combine the high-level information obtained with the encoder with the detailed spatial information before downsampling. While retaining the original concepts from the U-Net architecture, various U-Net architectures have been proposed to enhance its performance and adapt it to different tasks \cite{unet-attention, r2unet, dense-unet}. 

We propose the simple 2D U-Net instead of the more computational complex 3D U-Net baseline provided. In the 2D U-Net, which inherently lacks the ability to directly manage time, time is processed by extending the number of channels for the filters in the input and output layers of the U-Net to include the temporal dimension. This means that time is stacked on top of channels, so for an input with dimensions T x C x H x W, the input to the 2D U-Net architecture becomes T*C x H x W. Similarly, the output of the 2D U-Net of size T'*C' x H x W is unstacked to T' x C' x H x W, where T' is the forecasted timesteps and C' the output channels for each timestep. Whereas the 3D U-Net preserves the temporal dimension throughout the model, the 2D U-Net only combines information across time at the input layer, yet still manages to improve performance.

\subsubsection{ResNet 2D U-Net}
\label{s:unet}
One alternative to the original 2D U-Net consists of using ResNet blocks as the encoder to address the vanishing gradient problem in deep neural networks. ResNet employs residual connections or "identity shortcut" connections to skip one or more layers, providing an alternative shortcut path for gradient flow from later layers to earlier layers \cite{resnet}.

By replacing the U-Net encoder with a ResNet-18 encoder, the architecture is able to learn more complex features while still training efficiently with the aid of residual connections. Note that skip connections and residual connections are similar, but U-Net uses skip connections to make upsampling better, while ResNet uses residual connections for gradient flow. In the proposed architecture, every encoder block has a corresponding decoder block that processes the concatenation of the feature map coming from the encoder via skip connection and the upsampled feature map from the previous block in the decoder.

\subsubsection{Swin 2D U-Net}
\label{s:swin}
We also study the applications of Transformers, specifically Swin Transformer \cite{liu_swin_2021}, as the encoder. This choice is motivated by the Swin Transformer's capability to compute attention, facilitating the modeling of long-range dependencies to generate accurate forecasts. Notably, the Swin Transformer enhances the original Vision Transformer \cite{dosovitskiy_image_2021} by introducing Shifted Windows, enabling global attention with local computations in non-overlapping windows.

The Swin Transformer constructs a hierarchical representation by starting with small-sized patches and progressively merging neighboring patches in deeper Transformer layers. It computes multi-head self-attention \cite{vaswani_attention_2023} locally within non-overlapping windows that partition an image with a fixed number of patches per window. The Swin Transformer also introduces relative position bias in the attention computation for each head based on the spatial positions of each token.

The hierarchical representation obtained with the Swin Transformer is used by the decoder and skip connections to upsample the encoded feature maps and generate the required output.

\subsection{Dataset Optimization}
\label{s:data}
We further improve model performance and training time by maximizing the impact of training data. We use sampling techniques and adapt input data by selecting the appropriate input features and context for the problem.

\subsubsection{Data Sampling}
Sampling is essential as the dataset is large and ground-radar observations mostly contain regions with little or no precipitation that heavily bias the model towards predicting no rain. \textit{Importance sampling} \cite{sampling} is used to create a balanced dataset more representative of high precipitation by reducing the number of samples containing no precipitation. This optimization allows the model to train faster and improve performance with observations that positively help the model to learn from rainfall events.

Importance sampling computes $q_n$ for every sample as the acceptance probability for a given sample to be included into the subset of samples. This is calculated as 
$q_n = min{(1, q_{min} + x_n^{sat})}$, 
with $x_n^{sat}$ being the average of the saturated values over the output region $\frac{1}{C} \sum_c x_{n,c}^{sat}$. Note that $x_{n,c}^{sat}$ is computed as $1-exp(-x_{n,c})$ for every time step and pixel on the patch region. $q_{min}$, the minimum probability of inclusion even if there is not any rainfall, is set to $1x10^{-6}$ to obtain more events with high precipitation \cite{ravuri_skilful_2021}. The motivation for these formulas is to guarantee a minimum probability of inclusion for patches without significant precipitation, while still favouring patches with significant rain.

\subsubsection{Data Preparation}
In addition to the sampling process, we tailor the dataset to address the presented precipitation nowcasting problem with its requirements and available data. Consequently, the provided code is modified to load static data along with each sample, incorporating latitude, longitude, and topological height for each position within the input patch. This additional information holds the potential to impact the models by introducing relevant data that influences weather conditions, as discussed in previous studies \cite{bi_pangu-weather_2022, metnet3}.

However, it is not only important during data preparation for precipitation nowcasting to consider the features to be included in the input, but also the context provided. While larger context can enrich the information available for the models to learn from, it may pose challenges due to the increased dimensionality of the data, significantly affecting computation requirements and training time. For this specific problem, in terms of temporal context, only four timesteps representing the previous hour are provided and remain unmodified. However, as detailed in Section \ref{competition}, the spatial context provided is sufficient for generating forecasts with a lead time of up to 15-19 hours. Given that the competition focuses on forecasts up to 8 hours ahead, the spatial context of the input is reduced to optimize for training time and facilitate learning without introducing unnecessary data. As a result, the input patches are cropped from the original 252 x 252 pixels to 128 x 128 pixels, providing a spatial context of 516 km in each direction, which we find sufficient for the 8-hour prediction.

\subsection{Cross Entropy Loss}
\label{s:loss}
While the initially provided loss function is Mean Squared Error (MSE), a regression loss, classification losses such as cross-entropy have been used for precipitation nowcasting to obtain better performance \cite{metnet, metnet2, metnet3} and generate probabilistic forecasts that provide more information. For example, a summer shower, predicted with a 100\% probability, may produce 2mm of rain, while a thunderstorm generating 10mm of rain might only have a 20\% probability. Despite both events resulting in the same 2mm of rain per hour, the appropriate actions taken in response to each would differ significantly. Thus, probabilistic forecasts are more meaningful than intensity predictions.

Approaching precipitation nowcasting as a classification problem requires the use of predefined classes or buckets to categorize different levels of rain intensities. The classes utilized align with those presented in Table \ref{val-data-distribution}, aiming to establish narrower ranges for lower precipitation values where observations are more abundant, while still being capable of representing instances of heavy precipitation.

Another advantage of targeting precipitation nowcasting as a classification problem is that weights can be assigned to the different classes to deal with unbalanced datasets. The weights are set to be inversely proportional to frequency of classes, so that less frequent classes (those with high precipitation events) get higher weight than those that appear often (no rain or low rain).

While probabilistic outputs are more informative, forecast intensities may be required for some use cases, such as submitting to the leaderboard or evaluating against some metrics. In the case of cross-entropy, where probabilities are allocated across different classes that sum up to 1, the average of each bucket is multiplied by its probability, and the resulting values are summed to yield the final precipitation value. However, it should be noted that the probability distribution itself provides more informative insights than solely the computed value.

\subsection{Conditioning Lead Time}
\label{s:clt}
Precipitation nowcasting models can be designed to output predictions for all time steps in a single iteration or one at a time. While the initial approach produces all precipitation forecasts for the required lead times at once, the concept of conditioning lead time introduced in the MetNet \cite{metnet} architecture is studied. 

With conditioning lead time, a forward pass of the model makes a prediction for a single lead time. The required single lead time is provided as input to the model to condition every aspect of the computation. While in MetNet lead time is provided as a one-hot encoding, we provide it as a single value. Therefore, when using conditioning lead time, the model includes an additional channel as input that is entirely filled with the required timestep.

It is essential to note that this approach differs from autoregressive methods, where the model consistently predicts the next timestep using previously predicted timesteps as input. In our scenario, the model does not utilize any predictions as inputs and can generate a forecast for the desired lead time independently and simultaneously. Designing models able to forecast multiple timesteps yields improved results because autoregressive approaches incorporate both past observations and predicted forecasts, causing errors in the intermediate forecasts to accumulate and impact future predictions that rely on them as input.

\subsection{Crop \& Upsample}
\label{s:upsample}
The output of the 2D U-Net architecture has the same spatial dimensions as its input. This means that for an input sequence of size 128 x 128 pixels, a forward pass through the U-Net will generate an output of size 128 x 128 pixels. However, the labels correspond to the central patch of size 42 x 42 pixels. Therefore, to guide the precipitation nowcasting model, we take the central 42 x 42 pixels patch and upsample to the 252 x 252 pixels label. This cropping and upsample was introduced in MetNet \cite{metnet2} and is due to the different spatial resolutions of the input and labels, as explained in Section \ref{competition}.

An alternative would be to let the output of the U-Net be the forecast for the label region without cropping and upsampling the data. However, this would require the U-Net to learn how to obtain a high resolution output of the central region of the input region, which is significantly more challenging.

For upsampling, different methods are studied, including standard upsample algorithms such as nearest neighbor and bilinear interpolation, but also more complex super-resolution architectures such as Enhanced Deep Super Resolution (EDSR) \cite{lim_enhanced_2017} and NinaSR \cite{gouvine_ninasr_2021}. Nearest neighbor upsample copies the value from the nearest pixel, and bilinear uses all nearby pixels to calculate the pixel's value using linear interpolations. Both super-resolution architectures, EDSR and NinaSR, use residual blocks to improve over simple CNN techniques by making connections between the early layers of the network and the late ones, achieving good performance in terms of evaluation metrics and training time.

\section{Experimental Evaluation}
All models have been trained on Amazon AWS, utilizing a single Amazon EC2 G5 instance equipped with an NVIDIA A10G Tensor Core GPU. Table \ref{scores} summarizes the experiments conducted for the respective leaderboards, showcasing the achieved score, training time, and a brief description. Additional information on the configuration of the experiments can be found in Table \ref{experiments} in the Appendix.

\begin{table}[h]
  \caption{Experiments with obtained scored on the respective leaderboard.}
  \label{scores}
  \centering
  \begin{tabular}{lllll}
    \toprule
    Leaderboard & Experiment & Score & Time (h) & Comments \\
    \midrule
    \textit{Core} & \textit{Official Baseline} & \textit{0.0444} & \textit{-} & \textit{Official Baseline based on 3D U-Net} \\
    Core & exp1 & 0.0306 & 11.5 & ResNet 2D U-Net \& Cross-Entropy \\
    Core & exp2 & 0.0491 & 1.7 & exp1 \& sampling \\
    Core & exp3 & 0.0345 & 1.7 & exp2 \& buckets for CSI thresholds \\
    Core & exp4 & 0.0502 & 1.5 & exp2 \& crop and bilinear upsample \\
    Core & exp5 & 0.0477 & 2.5 & exp4 \& Conditioning Lead Time \\
    Core & exp6 & 0.0451 & 1.1 & exp4 \& class weights \\
    Core & exp7 & 0.0482 & 2.5 & exp4 \& EDSR upsample \\
    Core & exp8 & \textbf{0.0507} & 3.2 & exp4 \& NinaSR upsample \\
    \midrule
    \textit{Nowcasting} & \textit{Official Baseline} & \textit{0.0482} & - & \textit{Official   Baseline based on 3D U-Net} \\
    Nowcasting & exp1 & \textbf{0.0585} & 3.2 & exp8 from \textit{Core} (16 first   predictions) \\
    Nowcasting & exp2 & 0.0580 & 0.8 & U-Net for 4h \\
    Nowcasting & exp3 & 0.0552 & 2.0 & Swin for 4h \\
    \midrule
    \textit{Transfer} & \textit{Official Baseline} & \textit{0.0459} & \textit{-} & \textit{Official Baseline based on 3D U-Net} \\
    Transfer & exp1 & 0.0566 & 3.2 & exp8 from core (16 first predictions) \\
    Transfer & exp2 & \textbf{0.0584} & 0.8 & exp2 from \textit{Nowcasting} \\
    Transfer & exp3 & 0.0531 & 2.0 & exp3 from \textit{Nowcasting} \\
    \bottomrule
  \end{tabular}
\end{table}

The best results for the core challenge are achieved with experiment \#8, which uses the ResNet-18 2D U-Net architecture, the optimized dataset, cross-entropy loss function, and cropping and upsampling with NinaSR. This particular model is also utilized for the 4-hour prediction in both the nowcasting and transfer learning leaderboards (exp1), focusing solely on the initial 16 predicted timesteps. Remarkably, this model excels in the nowcasting leaderboard, outperforming other models trained exclusively for a 4-hour prediction. However, in the transfer learning leaderboard, this is not the case and training a new model for the 4-hour prediction appears to yield superior results indicating that for the transfer learning task, the model does not benefit from observing longer lead times.

\begin{figure}[h]
    \centering
    \includegraphics[width=0.5\textwidth]{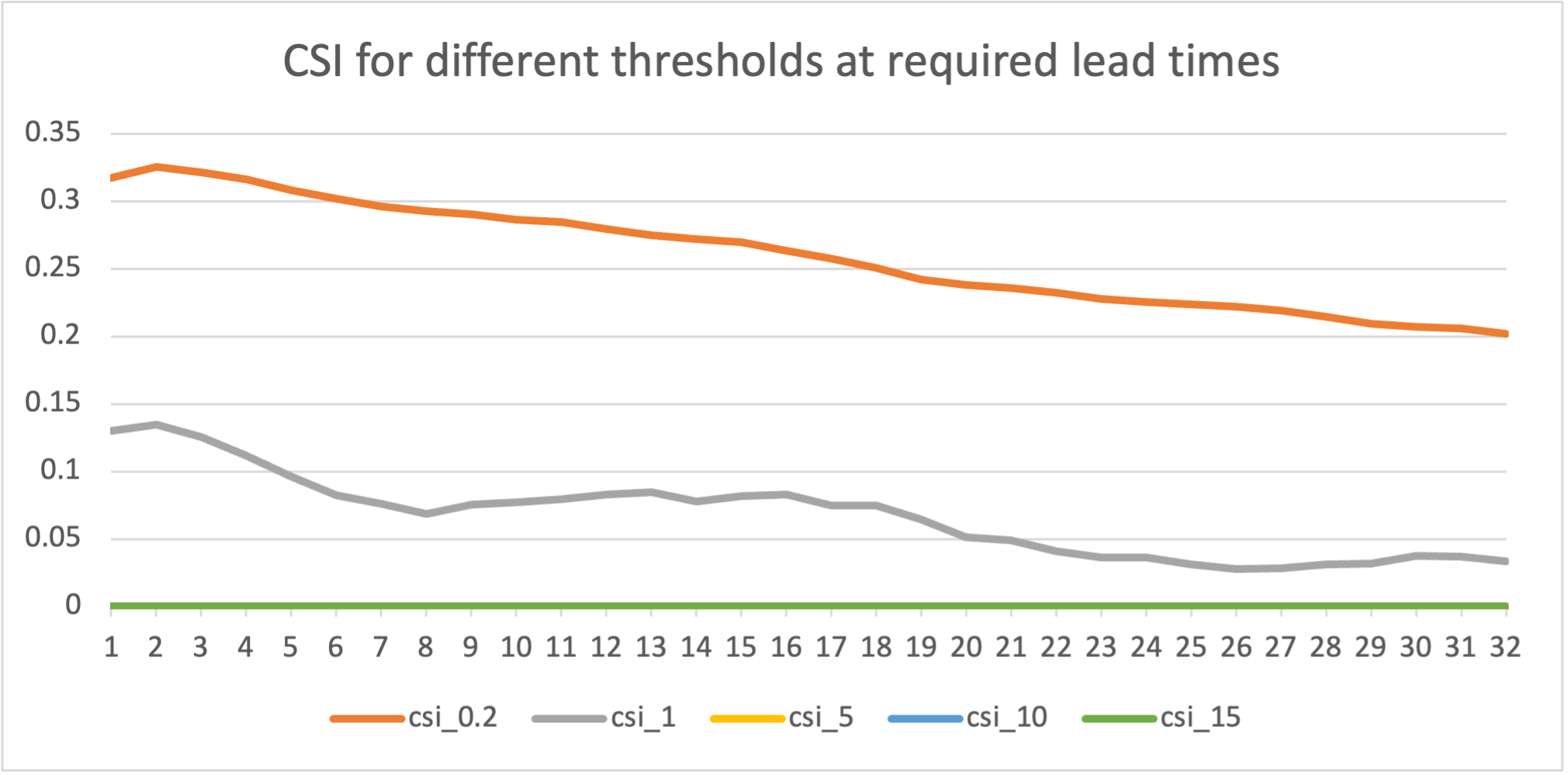}
    \caption{Evolution of Critical Success Index (CSI) at different thresholds for the required lead times for the core challenge (exp4).}
    \label{fig:test4-lt}
\end{figure}

As anticipated, the visualization of the Critical Success Index (CSI) across the various thresholds of the leaderboard at the required lead times, as depicted in Figure \ref{fig:test4-lt}, reveals a decrease in values for longer lead times due to increased uncertainty and complexity. Nevertheless, it is essential to note that throughout the experiments, the CSI values for thresholds above 5mm/h consistently score 0, indicating a general inability to predict these high-precipitation events, likely due to the under-represented examples with more than 5mm/h as shown in Table \ref{val-data-distribution}.

\begin{figure}[h]
    \centering
    \includegraphics[width=\textwidth]{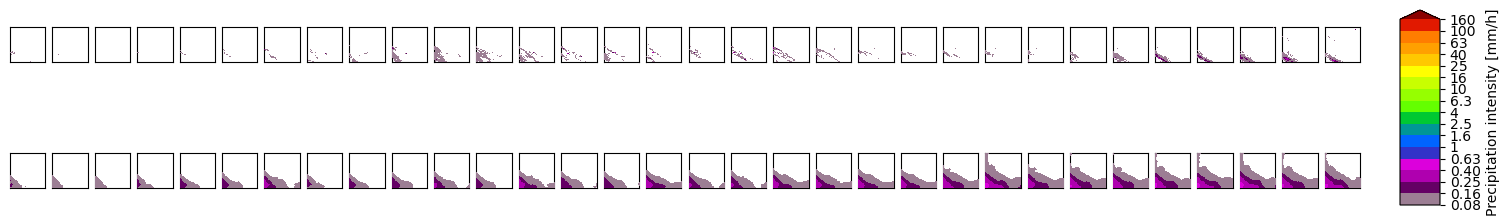}
    \caption{Visualization of ground truth (top) and predicted precipitation (bottom) for the required lead times for the core challenge for a validation sequence of the provided dataset (exp4).}
    \label{fig:test4}
\end{figure}

In Figure \ref{fig:test4}, ground truth and predictions for a validation sequence using the model from experiment \#4 are presented. This experiment is the same as experiment \#8 just explained, with the only change being that upsampling is performed via bilinear interpolation, so it was faster to train with minimal decrease in evaluation metrics. Please note that the precipitation in the bottom left corner is detected, but the resolution and intensity are not as high as in the ground truth. This is one of the problems that has been identified, most likely due to the lack of high-resolution input data.

\section{Conclusion \& Future Work}
The presented work is a part of the Weather4Cast 2023 NeurIPS competition, focusing on precipitation nowcasting using low-resolution input satellite observations. This challenge requires the application of data fusion, multi-channel video frame prediction, and super-resolution techniques.

The proposed models leverage the 2D U-Net architecture, incorporating either ResNet-18 or Swin Transformer blocks, an optimized dataset, cross-entropy loss function, conditioning lead time, and upsampling methods. The best performing models use the ResNet-18 2D U-Net architecture, the optimized dataset, cross entropy loss, and NinaSR upsample. However, similar performance is achieved with bilinear interpolation and it is faster to train. While not all these optimizations are necessarily employed by the best-performing models, specifically the Swin Transformers as the encoder or conditioning lead time, experiments that incorporate these enhancements consistently outperform the official baseline on the leaderboard in terms of evaluation metrics.

Moreover, this challenge highlights the complexity of successfully performing precipitation nowcasting from low-resolution satellite observations, particularly to anticipate high-precipitation events.

A primary area of concern that we will aim to address in future work is enhancing the accuracy when forecasting high-precipitation events. We believe that adding specific modules with different loss functions to the architecture could assist in identifying and predicting these instances of high precipitation. These new modules can guide the architecture to learn in an easier way, similar to the Crop \& Upsample optimization. Regarding loss functions, a limitation of cross-entropy is that it assigns the same loss for incorrect predictions regardless of the difference between the prediction and the ground truth.

In order to improve the overall performance, MetNet-2 \cite{metnet2} and MetNet-3 \cite{metnet3} propose optimizing the conditioning lead time strategy. Instead of merely having the required timestep as input to the model, this optimization entails influencing computation at every layer based on the lead time with a bias and scale vector derived from the required timestep.

Finally, another focus area is processing the temporal dimension of the input sequences to learn more intricate precipitation patterns and enhance overall performance by utilizing memory-based architectures or architectures that independently process the temporal dimension (i.e., Video Swin Transformer \cite{liu_video_2021}).

\newpage
\section*{Appendix}
Table \ref{experiments} provides a breakdown of the experiments conducted for each leaderboard in the Weather4Cast 2023 competition. In the table, RU and SU refer to ResNet 2D U-Net and Swin 2D U-Net models, respectively. CLT, LR, and BS correspond to conditioning lead time, learning rate, and batch size, with batch size indicating both the size that fits in memory and the effective batch size for gradient updates. Lead Time, LT, represents the maximum number of timesteps for which the model is designed to make predictions. Cross-entropy loss function is utilized for the training of all models, and static data is incorporated in every experiment except for experiment \#1.

\begin{table}[h]
  \caption{Details for the experiments run to train the different models.}
  \label{experiments}
  \centering
  \begin{tabular}{llllllllllll}
    Experiment & Model & Samples & Ep & Buckets & Up & CLT & LR & BS & LT \\
    \midrule
    Core exp1 & RU & 220K & 1 & mmh & No & No & 1E-4 & 16 (16) & 32 \\
    Core exp2 & RU & 15K & 4 & mmh & No & No & 1E-4 & 16 (16) & 32 \\
    Core exp3 & RU & 15K & 4 & w4c23\_1 & No & No & 1E-4 & 16 (16) & 32 \\
    Core exp4 & RU & 15K & 4 & mmh & Bilinear & No & 1E-4 & 8 (16) & 32 \\
    Core exp5 & RU & 15K & 3 & mmh & Bilinear & Yes & 1E-4 & 2 (32) & 32 \\
    Core exp6 & RU & 10K & 4 & mmh\_w & Bilinear & No & 1E-3 & 16 (16) & 32 \\
    Core exp7 & RU & 15K & 4 & mmh & EDSR & No & 1E-4 & 8 (16) & 32 \\
    Core exp8 & RU & 15K & 4 & mmh & NinaSR & No & 1E-4 & 8 (16) & 32 \\
    \midrule
    Nowcast exp1 & RU & 15K & 4 & mmh & NinaSR & No & 1E-4 & 8 (16) & 32 \\
    Nowcast exp2 & RU & 10K & 13 & mmh & Nearest & No & 1E-4 & 4 (32) & 16 \\
    Nowcast exp3 & SU & 2K & 19 & w4c23\_2 & Nearest & Yes & 1E-4 & 2 (128) & 16 \\
    \midrule
    Transfer exp1 & RU & 15K & 4 & mmh & NinaSR & No & 1E-4 & 8 (16) & 32 \\
    Transfer exp2 & RU & 10K & 13 & mmh & Nearest & No & 1E-4 & 4 (32) & 16 \\
    Transfer exp3 & SU & 2K & 19 & w4c23\_2 & Nearest & Yes & 1E-4 & 2 (128) & 16 \\
    \bottomrule
  \end{tabular}
\end{table}


\printbibliography

\end{document}